\documentclass{article}

     \PassOptionsToPackage{square,sort,comma,numbers}{natbib}
    \usepackage[preprint]{neurips_2019}

\usepackage[]{neurips_2019}

\usepackage[utf8]{inputenc} % allow utf-8 input
\usepackage[T1]{fontenc}    % use 8-bit T1 fonts
\usepackage{hyperref}       % hyperlinks
\usepackage{url}            % simple URL typesetting
\usepackage{booktabs}       % professional-quality tables
\usepackage{amsfonts}       % blackboard math symbols
\usepackage{nicefrac}       % compact symbols for 1/2, etc.
\usepackage{microtype}      % microtypography

\usepackage{graphicx}
\usepackage{algorithm,algpseudocode}
\usepackage{adjustbox}
\newtheorem{exmp}{Example}
\algnewcommand\algorithmicinput{\textbf{Input:}}
\algnewcommand\Input{\item[\algorithmicinput]}
\algnewcommand\algorithmicoutput{\textbf{Output:}}
\algnewcommand\Output{\item[\algorithmicoutput]}
\algnewcommand\algorithmicforeach{\textbf{for each}}
\algdef{S}[FOR]{ForEach}[1]{\algorithmicforeach\ #1\ \algorithmicdo}

\title{Induction of Non-Monotonic Rules From Statistical Learning Models Using High-Utility Itemset Mining}

\author{%
  Farhad Shakerin \\
  Department of Computer Science\\
  The University of Texas at Dallas\\
  Richardson, TX 75080 \\
  \texttt{farhad.shakerin@utdallas.edu} \\
  \And
   Gopal Gupta\\
   Department of Computer Science\\
   The University of Texas at Dallas\\
   Richardson, TX 75080 \\
   \texttt{gupta@utdallas.edu}
}

\begin{document}

\maketitle

\begin{abstract}
We present a fast and scalable algorithm to induce \textit{non-monotonic} logic programs from statistical learning models. We reduce the problem of search for best clauses to instances of the \textit{High-Utility Itemset Mining} (HUIM) problem. In the HUIM problem, feature values and their importance are treated as transactions and utilities respectively. We make use of TreeExplainer, a fast and scalable implementation of the Explainable AI tool SHAP, to extract locally important features and their weights from ensemble tree models. Our experiments with UCI standard benchmarks suggest a significant improvement in terms of classification evaluation metrics and running time of the training algorithm compared to ALEPH, a state-of-the-art \textit{Inductive Logic Programming} (ILP) system.
 \end{abstract}

\section{Introduction}
Dramatic success of machine learning has led to a torrent of Artificial Intelligence (AI) applications. However, the effectiveness of these systems is limited by the machines' current inability to explain their decisions and actions to human users. That's mainly because the statistical machine learning methods produce models that are complex algebraic solutions to optimization problems such as risk minimization
or geometric margin maximization. Lack of intuitive descriptions makes it hard for users to understand and verify the underlying rules that govern the model. Also, these methods cannot produce a justification for a prediction they arrive at for a new data sample.

The Explainable AI program \cite{xai} aims to create a suite of machine learning techniques that: a) Produce more explainable models, while maintaining a high level of prediction accuracy. b) Enable human users to understand, appropriately trust, and effectively manage the emerging generation of artificially intelligent partners. 

Inductive Logic Programming (ILP) \cite{ilp} is one Machine Learning technique where the learned model is in the form of logic programming rules (Horn Clauses) that are comprehensible to humans. It allows the background knowledge to be incrementally extended without requiring the entire model to be re-learned. Meanwhile, the comprehensibility of symbolic rules makes it easier for users to understand and verify induced models and even refine them.

The ILP learning problem can be regarded as a search problem for a set of clauses that deduce the training examples. The search is performed either top down or bottom-up. A bottom-up approach builds most-specific clauses from the training examples and searches the hypothesis space by using generalization. This approach is not applicable to large-scale datasets, nor it can incorporate \textit{negation-as-failure} into the hypotheses. A survey of bottom-up ILP systems and their shortcomings can be found at \cite{sakama05}. In contrast, top-down approach starts with the most general clause and then specializes it. A top-down algorithm guided by heuristics is better suited for large-scale and/or noisy datasets \cite{quickfoil}.

\begin{figure}
\centering
    \includegraphics[width=0.48\textwidth,scale=0.2]{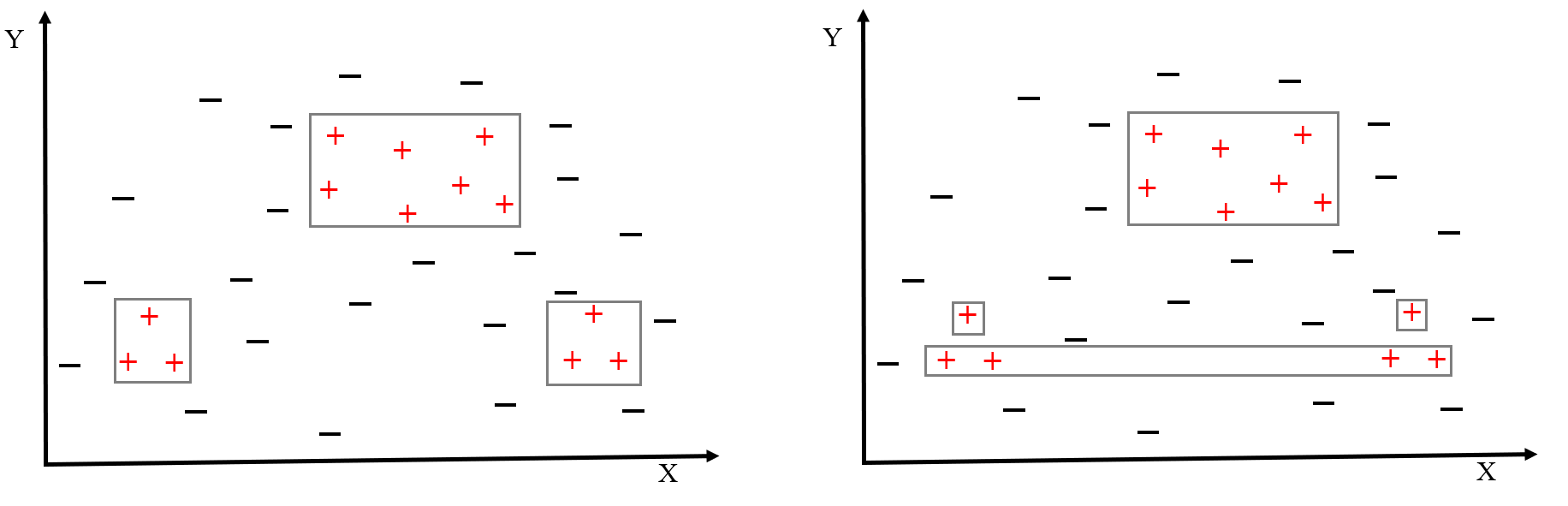}
\caption{Optimal sequential covering with 3 Clauses (Left), Sub-Optimal sequential covering with 4 Clauses (Right)}
\label{fig:localoptima}
\end{figure}

The FOIL algorithm by Quinlan \cite{foil} is a popular top-down algorithm. FOIL uses  heuristics from information theory called \textit{weighted information gain}. The use of a greedy heuristic allows FOIL to run much faster than bottom-up approaches and scale up much better. However, scalability comes at the expense of losing accuracy if the algorithm is stuck in a local optima and/or when the number of examples is insufficient. Figure \ref{fig:localoptima} demonstrates how the local optima results in discovering sub-optimal rules that does not necessarily coincide with the real underlying sub-concepts of the data. Additionally, since the objective is to learn pure clauses (i.e., clauses with zero or few negative example coverage) the FOIL algorithm often discovers too many clauses each of which only cover a few examples. Discovery of a huge number of clauses reduces the interpretability and also it does not generalize well on the test data.

Unlike top-down ILP algorithms, statistical machine learning algorithms are bound to find the relevant features because they optimize an objective function with respect to global constraints. This results in models that are inherently complex and cannot explain what features account for a classification decision on any given data sample. The Explainable AI techniques such as LIME \cite{lime} and SHAP \cite{shap} have been proposed that provide explanations  for any given data sample. Each explanation is a set of feature-value pairs that would locally determine what features and how strongly each feature, relative to other features, contributes to the classification decision. To capture the global behavior of a black-box model, however, an algorithm needs to group similar data samples (i.e., data samples for which the same set of feature values are responsible for the choice of classification) and cover them with the same clause. While in FOIL, the search for a clause is guided by heuristics, in our novel approach, we adapt \textit{High Utility Item-set Mining} (HUIM) \cite{huim} --- a popular technique from data mining --- to find clauses. We call this algorithm SHAP-FOLD from here on. The advantage of SHAP-FOLD over heuristics-based algorithms such as FOIL is that:
\begin{enumerate}
    \item SHAP-FOLD does not get stuck in a local optima
    \item SHAP-FOLD distinguishes exceptional cases from noisy samples
    \item SHAP-FOLD learns a reasonable number of non-monotonic rules in the form of default theories that would understandably capture the global behavior of any black-box model
    \item SHAP-FOLD is fast and scalable compared to conventional ILP algorithms
\end{enumerate}
This paper makes the following novel contribution: We
present a new ILP algorithm capable of learning \textit{non-monotonic}
logic programs from local explanations of
black-box models provided by SHAP. Our experiments on UCI standard benchmark data sets suggest that SHAP-FOLD outperforms ALEPH
in terms of classification evaluation metrics, running time, and providing more concise explanations measured in terms of number of clauses induced.
\iffalse
The rest of this paper is organized as follows: In Section \ref{sec:background} we present the problem definition as well as the background material to understand the SHAP-FOLD algorithm. In Section \ref{sec:SHAP-FOLD} we present the SHAP-FOLD algorithm. In Section \ref{sec:experiments} we present our experiments on UCI benchmark datasets. In Section \ref{sec:relatedworks} we discuss existing approaches and compare them with our SHAP-FOLD algorithm, and finally, in Section \ref{sec:conclusion} we present our conclusions with recommendations for future research.
\fi

\section{Background}
\label{sec:background}
\subsection{Inductive Logic Programming}
Inductive Logic Programming (ILP) \cite{ilp} is a subfield of machine learning that learns models in the form of logic programming rules (Horn Clauses) that are comprehensible to humans. This problem is formally defined as:\\
\textbf{Given}
\begin{enumerate}
    \item a background theory $B$, in the form of an extended logic program, i.e., clauses of the form $h \leftarrow l_1, ... , l_m,\ not \ l_{m+1},...,\ not \ l_n$, where $l_1,...,l_n$ are positive literals and \textit{not} denotes \textit{negation-as-failure} (NAF) \cite{Baral} and $B$ has no even cycle
    \item two disjoint sets of ground target predicates $E^+, E^-$ known as positive and negative examples respectively
    \item a hypothesis language of function free predicates $L$, and a  refinement operator $\rho$ under $\theta-subsumption$ \cite{plotkin70} that would disallow even cycles.
\end{enumerate}
\textbf{Find} a set of clauses $H$ such that:
\begin{itemize}
    \item $ \forall e \in \ E^+ ,\  B \cup H \models e$
    \item $ \forall e \in \ E^- ,\  B \cup H \not \models e$
    \item $B \land H$ is consistent.
\end{itemize}
\subsection{The FOIL Algorithm}
FOIL is a top-down ILP algorithm that follows a \textit{sequential covering} scheme to induce a hypotheses. The FOIL algorithm is summarized in Algorithm \ref{algo:foil}. This algorithm repeatedly searches for clauses that score best with respect to a subset of 
positive and negative examples, a current hypothesis and a heuristic called \textit{information gain} (IG). 

\begin{algorithm}
\caption{Summarizing the FOIL algorithm}
\label{algo:foil}
\begin{algorithmic}[1]
\Input $target,B,E^+,E^-$ 
\Output 
Initialize $H \gets \emptyset $
\While{($|E^+| > 0$)}
	\State $c \gets (target$ :- $ \ true.)$
	\While{($|E^-| > 0 \land c.length < max\_length $)}
		\For{all $ \ c' \in \rho (c)$}
        	\State $compute \ score(E^+,E^-,H \cup \{c'\},B)$
    	\EndFor
    	\State let $\hat{c}$ be the $c' \in \rho(c)$ with the best score   
         \State $E^- \gets covers(\hat{c},E^-)$
    \EndWhile	
    \State add $\hat{c}$ to $H$
    \State $E^+ \gets E^+ \setminus covers(\hat{c},E^+)$
\EndWhile 
\State \textbf{return} $H$
\end{algorithmic}
\end{algorithm}

The inner loop searches for a clause with the highest information gain using a general-to-specific hill-climbing search. To specialize a given clause $c$, a refinement operator $\rho$ under $\theta$-subsumption \cite{plotkin70} is employed. The most general clause is $p(X_1,...,X_n) \gets true.$ where the predicate $p/n$ is the predicate being learned and each $X_i$ is a variable. The refinement operator specializes the current clause $h \gets b_1,...b_n .$ This is realized by adding a new literal $l$ to the clause yielding $h \gets b_1,...b_n,l$. The heuristic based search uses information gain.

\subsection{SHAP}
SHAP \cite{shap} (SHapley Additive exPlanations) is a unified approach with foundations in game theory to explain the output of any machine learning model in terms of its features' contributions. To compute each feature $i$'s contribution, SHAP requires retraining the model on all feature subsets $S \subseteq F$, where $F$ is the set of all features. For any feature $i$, a model $f_{S \cup \{i\}}$ is trained with the feature $i$ present, and another model $f_S$ is trained with feature $i$ eliminated. Then, the difference between predictions is computed as follows: $f_{S \cup \{i\}}(x_{S \cup \{i\}}) - f_S(x_S)$, where $x_S$ represents sample's feature values in $S$. Since the effect of withholding a feature depends on other features in the model, the above differences are computed for all possible subsets of $S \subseteq F \setminus \{i\}$ and their average taken. The weighted average of all possible differences (a.k.a Shapley value) is used as feature importance. The Equation \ref{eq:shapleynumber} shows how Shapley value associated with each feature value is computed:
\begin{equation}
\label{eq:shapleynumber}
\phi_i = \sum_{S \subseteq F \setminus \{i\}} \frac{|S|!(|F| - |S| -1)!}{|F|!} \bigg [f_{S \cup \{i\}}(x_{S \cup \{i\}}) - f_S(x_S) \bigg ]
\end{equation}

\begin{figure}[h]
\centering
    \includegraphics[width=0.8\textwidth]{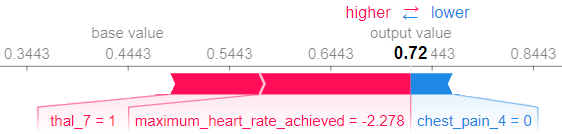}
\caption{Shap Values for A UCI Heart Prediction }
\label{fig:heartshap}
\end{figure}

 Given a dataset and a trained model, SHAP outputs a matrix with the shape $(\# samples, \#features)$ representing the Shapley value of each feature for each data sample. Each row sums to the difference between the model output for that sample and the expected value of the model output. This difference explains why the model is inclined to predict a specific class outcome.
\begin{exmp}
\label{ex:heart}
The UCI heart dataset contains features such as patient's  blood pressure, chest pain, thallium test results, number of major vessels blocked, etc. The classification task is to predict whether the subject suffers from heart disease or not. Figure \ref{fig:heartshap} shows how SHAP would explain a model's prediction over a data sample.   
\end{exmp}
For this individual, SHAP explains why the model predicts heart disease by returning the top features along with their Shapley values (importance weight). According to SHAP, the model predicts ``heart disease" because of the values of ``thalium test" and ``maximum heart rate achieved" which push the prediction from the base (expected) value of 0.44 towards a positive prediction (heart disease). On the other hand, the feature ``chest pain" would have pushed the prediction towards negative (healthy), but it is not strong enough to turn the prediction. 

The categorical features should be \textit{binarized} before any model is trained. Binarization (a.k.a one-hot encoding) is the process of transforming each categorical feature with domain of cardinality $N$, into $N$ new binary predicates (features).In Example \ref{ex:heart}, chest pain level is a categorical feature with 4 different values in the set $\{1,2,3,4\}$. Type 4 chest pain indicates asymptomatic pain and is a serious indication of a heart condition. In this case, binarization results in 4 different predicates. The ``thalium test" is also a categorical feature with outcomes in the set \{3,6,7\}. Any outcome other than 3, indicates a defect (6 for fixed and reversible for 7). In case of Example \ref{ex:heart}, SHAP determines that the feature ``thal\_7" with outcome of 1 (True), pushes the prediction towards heart disease. To reflect this fact in our ILP algorithm, for any person $X$, the predicate \texttt{thal(X,7)} is introduced. Also, SHAP indicates that the binary feature ``chest\_pain\_4" with value 0 (False), pushes the prediction towards healthy.

\subsection{High-Utility Itemset Mining}
The problem of \textit{High-Utility Itemset Mining} (HUIM) is an extension of an older problem in data mining known as \textit{frequent pattern mining} \cite{freqpm}. Frequent pattern mining is meant to find frequent patterns in transaction databases. A \textit{transaction database} is a set of records (transactions) indicating the items purchased by customers at different times. A \textit{frequent itemset} is a group of items that appear in many transactions. For instance, \{noodles, spicy sauce\} being a frequent itemset, can be used to take marketing decisions such as co-promoting noodles with spicy sauce. Finding frequent itemsets is a well-studied problem with an efficient algorithm named Apriori \cite{apriori}. However, in some applications frequency is not always the objective. For example, the pattern \{milk,bread\} may be highly frequent, but it may yield a low profit. On the other hand, a pattern such as \{caviar, champagne\} may not be frequent but may yield a high profit. Hence, to find interesting patterns in data, other aspects such as profitability is considered. 

Mining high utility itemsets can be viewd as a generalization of the frequent itemset mining where each item in each transaction has a utility (importance) associated with it and the goal is to find itemsets that generate high profit when for instance, they are sold together. The user has to provide a value for a threshold called \textit{minimum utility}. A high utility itemset mining algorithm outputs all the high-utility itemsets with at least \textit{minimum utility} profit. The HUIM problem is formally defined as follows:
\begin{itemize}
    \item $I = \{i_1, i_2,..., i_m\}$ is a set of items.
    \item $D = \{T_1, T_2, ..., T_n\}$ be a transaction database where each transaction $T_i \in D$ is a subset of $I$.
    \item $u(i_p,T_q)$ denotes the utility (profit) for item $i_p$ in transaction $T_q$. For example $u(b,T_0) = 10$ in Example from Table \ref{tbl:huiexample}.
    \item $u(X,T_q)$, utility of an itemset $X$ is defined as $\sum_{i_p \in X}^{} u(i_p,T_q)$, where $X = \{i_1, i_2, ..., i_K\}$ is a k-itemset, $X \subseteq T_q$ and $1 \leq K \leq m$.
    \item $u(X)$, utility of an itemset $X$ is defined as $\sum_{T_q \in D \land X \subseteq T_q} u(X,T_q)$.
\end{itemize}
An itemset $X$ is a \textit{high utility itemset} if $u(X) \geq \varepsilon$, where $X \subseteq I$ and $\varepsilon$ is the \textit{minimum utility} threshold, otherwise, it is a low utility itemset. Table \ref{tbl:huiexample} shows a transaction database consisting of 5 transactions. Left columns shows the transaction Identifier. Middle column contains the items included in each transaction and right column contains each item's respective profit. If the $\varepsilon$ is set to 25, the result of a high utility itemset mining algorithm is shown in the right table in Table \ref{tbl:huiexample}.

Several high utility itemset mining algorithms have been proposed (e.g., UMining, Two-Phase, IHUP, UP-Growth, etc). A complete survey of these algorithms can be found in \cite{huimsurvey}. The differences between these algorithms lies in the data structures, strategies that are employed for searching high utility itemsets (DFS vs. BFS), and how they prune the unpromising paths.

\begin{table}[h]
\centering
\begin{tabular}{|c|l|l|}
\hline
\multicolumn{1}{|l|}{\textbf{Transactions}} & \textbf{Items} & \textbf{Profits} \\ \hline
$\mathbf{T_0}$                              & a b c d e      & 5 10 1 6 3       \\ \hline
$\mathbf{T_1}$                              & b c d e        & 8 3 6 3          \\ \hline
$\mathbf{T_2}$                              & a c d          & 5 1 2            \\ \hline
$\mathbf{T_3}$                              & a c e          & 10 6 6           \\ \hline
$\mathbf{T_4}$                              & b c e          & 4 2 3            \\ \hline
\end{tabular}
\quad
\begin{tabular}{|l|l|}
\hline
\multicolumn{2}{|c|}{\textbf{High Utility Itemsets}} \\ \hline
\{a, c\}: 28           & \{a, c, e\}: 31    \\
\{a, b, c, d, e\}: 25  & \{b, c\}: 28       \\
\{b, c, d\}: 34        & \{b, c, d, e\}: 40 \\
\{b, c, e\}: 37        & \{b, d\}: 30       \\
\{b, d, e\}: 36        & \{b, e\}: 31       \\
\{c, e\}: 27           &                    \\ \hline
\end{tabular}
\caption{Left: A High Utility Itemset Problem Instance. Right: Solution for minutil = 25  }
\label{tbl:huiexample}
\end{table}

The downside of this approach is that it requires the decision maker to choose a \textit{minimum utility} threshold value for discovering interesting itemsets. This is quite challenging as too low a choice of $\varepsilon$ results in too many itemsets and too high a choice of $\varepsilon$ results in too few itemsets. In order to address this issue, Top-K High Utility Itemset (THUI) mining problem was introduced \cite{topk}, where the user wants to discover the $k$ itemsets having the highest utility. A top-k high-utility itemset mining algorithm works as follows: It initially sets an internal $\varepsilon$ threshold to 0, and starts to explore the search space. Then, as soon as k high utility itemsets are found, the internal $\varepsilon$ is raised to the utility of the pattern having the lowest utility among the current top-k patterns. Then, the search continues and for each high utility itemset found, the set
of the current top-k pattern is updated as well as the internal $\varepsilon$.
When the algorithm terminates, it returns the set of the top-k high utility itemsets.  

\section{SHAP-FOLD Algorithm}
\label{sec:SHAP-FOLD}
In this section we present the SHAP-FOLD algorithm. SHAP-FOLD learns a concept in terms of a default theory \cite{fold}. A default theory is a \textit{non-monotonic} logic theory to formalize reasoning with default assumptions in absence of complete information. In Logic Programming, default theories are represented using \textit{negation-as-failure} (NAF) semantics \cite{Baral}. 
\begin{exmp}
\label{ex:flies}
The following default theory ``Normally, birds fly except penguins which do not", is represented as:
\begin{verbatim}
    flies(X) :- bird(X), not ab_bird(X).
    ab_bird(X) :- penguin(X).
\end{verbatim}
This default theory is read as: ``For every object X, X flies if X is a bird and is not abnormal. For every object X, X is an abnormal bird if it is a penguin". 
\end{exmp}
The SHAP-FOLD algorithm adapts the FOIL style sequential covering scheme. Therefore, it iteratively learns single clauses, until all positive examples are covered. To learn one clause, SHAP-FOLD first finds common patterns among positive examples. If the resulted clause (default) covers a significant number of negative examples, SHAP-FOLD swaps the current positive and negative examples and recursively calls the algorithm to learn common patterns in negative examples (exceptions). As shown in Example \ref{ex:flies}, the exceptions are ruled out using negation-as-failure. Learning exceptions allow our SHAP-FOLD algorithm to distinguish between noisy samples and exceptional cases.

To search for ``best" clause, SHAP-FOLD tightly integrates the High Utility Itemset Mining (HUIM) and the SHAP technique. In this novel approach, the SHAP system is employed to find relevant features as well as their importance. To find the ``best" clause SHAP-FOLD creates instances of HUIM problem. Each instance, contains a subset of examples represented as a set of ``transactions" as shown in Table \ref{tbl:huiexample}. Each ``transaction" contains a subset of feature values along with their corresponding utility (i.e., feature importance). The feature importance $\phi_i \in [0,1]$ for all $i$ distinct feature values. Therefore, a \textit{high-utility itemset} in any set of ``transactions" represents strongest features that would contribute to the classification of a significant number of examples, because, otherwise, that itemset would not have been selected as a high-utility itemset. To find the itemset with highest utility, the HUIM algorithm Top-K \cite{topk} is invoked with $K$ set to 1. 

SHAP-FOLD takes a target predicate name (G), a tabular dataset (D) with $m$ rows and two different labels $+1$ and $-1$ for positive examples and negative examples respectively. $E^+$ and $E^-$ represent these examples in the form of target atoms. It also takes a ``transaction" database. Each row of T contains a subset of an example's feature-values ($\vec{z}_i$) along with their Shapley values ($\vec{\phi}_i$). This ``transaction" database is passed along to create HUIM instance and find the itemset with highest utility every time Top-K algorithm is invoked. The summary of SHAP-FOLD's pseudo-code is shown in Algorithm \ref{algo:SHAP-FOLD}.

In the function FOIL (lines 1-8), \textit{sequential covering } loop to cover positive examples is realized. On every iteration, a default clause (and possibly multiple exceptions) - denoted by $C_{def+exc}$ - is learned and added to the hypothesis. Then, the covered examples are removed from the remaining examples. In the function LEARN\_ONE\_RULE (lines 9-17), Top-K algorithm with $k=1$ is invoked and a high-utility itemset (i.e., a subset of features-values and their corresponding Shapley values) is retrieved. These subset of features create the default part of a new clause. Next, if the default clause covers false positives, the current positive and negative examples are swapped to learn exceptions. In the function LEARN\_EXCEPTIONS (lines 18 - 25), the algorithm recursively calls itself to learn clauses that would cover exceptional patterns. When the recursive call returns, for all learned clauses, their head is replaced by an abnormality predicate. To manufacture the complete default theory, the abnormality predicate preceded by negation-as-failure (not) is added to the default part. The following example shows how SHAP-FOLD learns a concise non-monotonic logic program from an XGBoost trained model. 
\begin{exmp}
\label{ex:car}
The ``UCI Cars" dataset has the information about evaluating 1728 different cars and their acceptability based on features such as buying price, maintenance cost, trunk size, capacity, number of doors, and safety. SHAP-FOLD generates the following program from a trained XGBoost model:   
\begin{verbatim}
 DEF(1):       acceptable(A):- safety(A,high), not ab0(A).
 EXCEPTIONS(1):       ab0(A):- persons(A,2).
                      ab0(A):- maintenance(A,very_high).
 DEF(2):       acceptable(A):- persons(A,4), safety(A,medium), not ab1(A).
 EXCEPTIONS(2):       ab1(A):- price(A,very_high), trunk(A,small).
                      ab1(A):- price(A,high), maintenance(A,very_high).
 DEF(3):       acceptable(A):- trunk(A,big),safety(A,medium), persons(A,>5).
\end{verbatim}
On first iteration, the clause DEF(1) (i.e., \texttt{acceptable(A) :- safety(A,high)} is generated. Since it covers a significant number of negative examples, $E^+$ and $E^-$ are swapped and algorithm recursively calls itself. Inside LEARN\_EXCEPTIONS, the recursive call returns with EXCEPTIONS(1) clauses. The head predicate \texttt{ab0} replaces their head and finally in line 24, the negation of abnormality is appended to the default to create a complete default clause. According to the discovered default clause, a car is considered acceptable if its safety is high, unless it can only fit two person (too small) or its maintenance cost is high. Similarly, the DEF(2) clause states that a car is acceptable if it can fit 4 person and its safety is medium, unless its price is too high and its trunk is small. Another exceptional case is established by high price and very high maintenance. The third clause default part does not cover any false positives, hence no exception clause is learned.  
\end{exmp}
There are some technicalities that should be pointed out: (1) The numeric features should be discretized (i.e., by splitting them into fixed number of intervals). This restriction is imposed by SHAP technique. (2) Inspired by FOIL implementation, the ``if statement" in line 13 of the algorithm is realized in terms of an empirical accuracy (e.g., \% 85) to avoid over-fitting. This would allow some noise error after learning the exceptions.    

\begin{algorithm}
\caption{Summary of SHAP-FOLD Algorithm}
\label{algo:SHAP-FOLD}
\begin{algorithmic}[1]
\Input  ~~~$G$: Target Predicate to Learn
\Statex ~~~~~~~~~$B$: Background Knowledge  
\Statex ~~~~~~~~~$D ~~= \{~ (\vec{x}_1,y_1),...,(\vec{x}_m,y_m) \}$ : ~$y_i \in \{-1,+1\}$
\Statex ~~~~~~~~~$E^+ = \{ ~\vec{x}_i ~|~ \vec{x}_i \in D \land y_i = 1\}$ :~ Positive Examples
\Statex ~~~~~~~~~$E^- = \{ ~\vec{x}_i ~|~ \vec{x}_i \in D \land y_i = -1\}$:~ Negative Examples
\Statex ~~~~~~~~~$T ~~~= \{~(\vec{z}_i, \vec{\phi}_i) ~|~ \vec{z}_i \subseteq \vec{x}_i \land \vec{x}_i \in D \land \vec{\phi}_i$ is $\vec{z}_i$'s Shapley values $\}$
\Output  $D ~~~= \{~C_1,...,C_n\}$ \Comment{default clauses}
\Statex ~~~~~~$AB ~~~= \{~ab_1,...,ab_m\}$ \Comment{exceptions/abnormal 
clauses}
\Function{FOIL}{$E^+,E^-$} 
\While{($|E^+| > 0$)}
\State $C_{def+exc} \gets$ \Call{Learn\_One\_Rule}{{$E^+$},{{$E^-$}}}
\State $E^+ \gets E^+ \setminus covers(C_{def+exc},E^+,B)$
\State $D \gets D \cup \{ C_{def+exc} \}$
\EndWhile
\State \textbf{return} $D, AB$ \Comment{returns sets of defaults and exceptions}
\EndFunction
\Function{Learn\_One\_Rule}{${E^+},{E^-}$}
\State  - let Item-Set be $\{(f_1,...f_n),(\phi_1,...,\phi_n)\} \gets$ \Call{Top-K}{K=1,{$E^+$},{T}}   \Comment{Call to HUIM algorithm}
\State $C_{def} \gets (G$ :- $f_1,...,f_n)$ 
\State $FP  \gets covers(C_{def},E^-)$ \Comment{FP denotes False Positives}
\If{$FP > 0$}
	\State $C_{def+exc} \gets \Call{LEARN\_EXCEPTIONS}{{C_{def}},{E^-},{{E^+}}}$
\EndIf
\State \textbf{return} $C_{def+exc}$ 
\EndFunction

\Function{LEARN\_EXCEPTIONS}{${C_{def}},{E^+},{E^-}$}
	\State $ \{C_1,...,C_k\} \gets \Call{FOIL}{E^+,E^-} $ \Comment{Recursive Call After Swapping}
	\State $ab\_index \gets GENERATE\_UNIQUE\_AB\_INDEX()$
	\For {$ i \gets 1$ \textbf{to} $k$}
		\State $AB \gets AB \cup \{ ab_{ab\_index} \ $:-$ \ bodyof(C_i) \}$
	\EndFor
	\State \textbf{return} $C_{def+exc} \gets (headof(C_{def}) \ $:-$ \ bodyof(C_{def}), \ \textbf{not}(ab_{ab\_index}))$

\EndFunction
\end{algorithmic}
\end{algorithm}

\section{Experiments}
\label{sec:experiments}
In this section, we present our experiments on UCI standard benchmarks \cite{uci}. The ALEPH system \cite{aleph} is used as the baseline. ALEPH is a state-of-the-art ILP system that has been widely used in prior work. To find a rule, ALEPH starts by building the most specific clause, which is called the ``bottom clause", that entails a seed example. Then, it uses a branch-and-bound algorithm to perform a general-to-specific heuristic search for a subset of literals from the bottom clause to form a more general rule. We set ALEPH to use the heuristic enumeration strategy, and the maximum number of branch nodes to be explored in a branch-and-bound search to 500K. We also configured Aleph to allow up to 50 false examples covered by each clause while each clause should be at least 80 \%  accurate. We use the standard metrics including precision, recall, accuracy and $F_1$ score to measure the quality of the results. 

The SHAP-FOLD requires a statistical model as input to the SHAP technique. While computing the Shapley values is slow, there is a fast and exact implementation called TreeExplainer \cite{treeshap} for ensemble tree models. XGBoost \cite{xgboost} is a powerful ensemble tree model that perfectly works with TreeExplainer. Thus, we trained an XGBoost model for each of the reported experiments in this paper. Table \ref{tbl:accuracies} presents the comparison between ALEPH and SHAP\_FOLD on classification evaluation of each UCI dataset. The best performer is highlighted with boldface font. In terms of the running time, SHAP-FOLD scales up much better. In case of ``King-Rook vs. King-Pawn", while ALEPH discovers 283 clauses in 836 seconds, SHAP-FOLD  does much better. It finishes in 8 seconds discovering only 3 clauses that cover the knowledge underlying the model. Similarly, in case of ``UCI kidney", SHAP-FOLD finds fewer significantly fewer clauses.Thus, not only SHAP-FOLD's performance is much better, it discovers more succinct programs. Also, scalability is a major problem in ILP, that our SHAP-FOLD algorithm solves: its execution performance is orders of magnitude better.
%GG: check above paragarph

\begin{table}[]
\begin{adjustbox}{max width=\textwidth}
\begin{tabular}{|l|c|c|c|c|c|c|c|c|c|c|c|}
\hline
           & \multicolumn{1}{l|}{} & \multicolumn{10}{c|}{Algorithm}                                                                                                                                                                                             \\ \hline
           & \multicolumn{1}{l|}{} & \multicolumn{5}{c|}{Aleph}                                                            & \multicolumn{5}{c|}{SHAP-FOLD}                                                                                                      \\ \hline
Data Set   & Shape                 & Precision                      & Recall & Accuracy & F1   & \multicolumn{1}{l|}{Time (s)} & Precision                      & Recall                         & Accuracy                       & F1   & \multicolumn{1}{l|}{Time (s)} \\ \hline
cars       & (1728, 6)             & 0.83                           & 0.63   & 0.85     & 0.72 & 73                        & \textbf{0.84} & \textbf{0.94} & \textbf{0.93} & \textbf{0.89} & \textbf{5}                         \\ \hline
credit-a   & (690, 15)             & 0.78                           & 0.72   & 0.78     & 0.75 & 180                       & \textbf{0.90} & \textbf{0.74} & \textbf{0.84} &\textbf{ 0.81} & \textbf{7}                         \\ \hline
breast-w   & (699, 9)              & 0.92                           & 0.87   & 0.93     & 0.89 & 10                        & \textbf{0.92} & \textbf{0.95} & \textbf{0.95} &\textbf{ 0.93} & \textbf{2}                         \\ \hline
kidney     & (400, 24)             & \textbf{0.96} & 0.92   & 0.93     & 0.94 & 5                         & 0.93                           & \textbf{0.95} & 0.93 & 0.94 & \textbf{1}                         \\ \hline
voting     & (435, 16)             & 0.97                           & 0.94   & 0.95     & 0.95 & 25                        & \textbf{0.98} & \textbf{0.98} & 0.95 & \textbf{0.96} & \textbf{1 }                        \\ \hline
autism     & (704, 17)             & 0.73                           & 0.43   & 0.79     & 0.53 & 476                       & \textbf{0.96} & \textbf{0.83} & \textbf{0.95} & \textbf{0.89} & \textbf{2}                         \\ \hline
ionosphere & (351, 34)             & \textbf{0.89} & 0.87   & 0.85     & 0.88 & 113                       & 0.87                           & \textbf{0.91} & 0.85 & \textbf{0.89} & \textbf{2}                         \\ \hline
heart      & (270, 13)             & 0.76                           & 0.75   & 0.78     & 0.75 & 28                        & 0.76 & \textbf{0.83} & \textbf{0.81} & \textbf{0.80} & \textbf{1}                         \\ \hline
kr vs. kp  & (3196, 36)            & 0.92           &  0.99  & 0.95      & 0.95  & 836                   & 0.92                           & 0.99                           & 0.95                           & 0.95 & \textbf{8}                         \\ \hline
\end{tabular}
\end{adjustbox}
\caption{Evaluation of SHAP\_FOLD on UCI Datasets}
\label{tbl:accuracies}
\end{table}
SHAP-FOLD almost always achieves a higher Recall score.This suggests that the proper use of \textit{negation-as-failure} leads to better coverage. The absence of negation from ALEPH hypothesis space forces the algorithm to create too specific clauses which leaves many positive examples uncovered. In contrast, our SHAP-FOLD algorithm emphasizes on better coverage via finding high-utility patterns of important features first. If the result turns out to cover too many negative examples to tolerate, by learning exceptions and ruling them out (via the same algorithm applied recursively), SHAP-FOLD maintains the same coverage as it rules out exceptional negative examples. 

SHAP-FOLD is a Java application that interfaces SWI-Prolog \cite{swiweb} using JPL library. The HUIM instances are solved by calling TKU from the SPMF Data mining library \cite{spmf}.

\section{Related Works and Conclusions}
\label{sec:relatedworks}
A survey of ILP can be found in \cite{ilp20}. In ILP community, researchers have tried to combine statistical methods with ILP techniques. Support Vector ILP \cite{svmilp} uses ILP hypotheses as kernel in dual form of the SVM algorithm. kFOIL \cite{kfoil} learns an incremental kernel for SVM algorithm using a FOIL style specialization. nFOIL \cite{nfoil} integrates the Naive-Bayes algorithm with FOIL. The advantage of our research over all of the above mentioned research work is that, first it is model agnostic, second it is scalable thanks to the fast and scalable HUIM algorithm and SHAP TreeExplainer, third it enjoys the power of \textit{negation-as-failure} which is absent from the above mentioned works.  

%\section{Conclusions}
%\label{sec:conclusion}
In this paper, we presented a fast and scalable ILP algorithm to induce default theories from statistical machine learning models. In this novel approach, irrelevant features are filtered out by SHAP, a technique from explainable AI. Then, the problem of searching for ``best" clause is reduced to a \textit{High-Utility Itemset Mining} problem. Our experiments on benchmark datasets suggest a significant improvement in terms of the classification evaluation metrics and running time. 

There are number of directions for future work: (i) Incorporating Relational features, and learning probabilistic logic programs that are currently absent from the research presented in this paper (ii) extending non-monotonic logic programs to image classification explanation domain. 

\subsubsection*{Acknowledgments}
Authors are partially supported by NSF Grant IIS 1718945.
We would like to cordially thank S. Mohammad Mirbagheri for bringing HUIM algorithms to our attention. We also thank the members of \textit{Applied Logic, Programming-languages \& Systems} (ALPS) lab for useful discussions and feedback.

\bibliographystyle{unsrt}
\bibliography{mycitations}

\begin{thebibliography}{10}

\bibitem{xai}
David Gunning.
\newblock Explainable artificial intelligence (xai),
  \url{https://www.darpa.mil/program/explainable-artificial-intelligence},
  2015.

\bibitem{ilp}
Stephen Muggleton.
\newblock Inductive logic programming.
\newblock {\em New Gen. Comput.}, 8(4):295--318, February 1991.

\bibitem{sakama05}
Chiaki Sakama.
\newblock Induction from answer sets in nonmonotonic logic programs.
\newblock {\em {ACM} Trans. Comput. Log.}, 6(2):203--231, 2005.

\bibitem{quickfoil}
Qiang Zeng, Jignesh~M. Patel, and David Page.
\newblock Quickfoil: Scalable inductive logic programming.
\newblock {\em Proc. VLDB Endow.}, 8(3):197--208, November 2014.

\bibitem{foil}
J.~Ross Quinlan.
\newblock Learning logical definitions from relations.
\newblock {\em Machine Learning}, 5:239--266, 1990.

\bibitem{lime}
Marco~Tulio Ribeiro, Sameer Singh, and Carlos Guestrin.
\newblock "why should {I} trust you?": Explaining the predictions of any
  classifier.
\newblock In {\em Proceedings of the 22nd {ACM} {SIGKDD} 2016}, pages
  1135--1144, 2016.

\bibitem{shap}
Scott~M Lundberg and Su-In Lee.
\newblock A unified approach to interpreting model predictions.
\newblock In {\em Advances in Neural Information Processing Systems}, pages
  4765--4774, 2017.

\bibitem{huim}
Wensheng Gan, Jerry~Chun{-}Wei Lin, Philippe Fournier{-}Viger, Han{-}Chieh
  Chao, Tzung{-}Pei Hong, and Hamido Fujita.
\newblock A survey of incremental high-utility itemset mining.
\newblock {\em Wiley Interdiscip. Rev. Data Min. Knowl. Discov.}, 8(2), 2018.

\bibitem{Baral}
Chitta Baral.
\newblock {\em Knowledge representation, reasoning and declarative problem
  solving}.
\newblock Cambridge University Press, Cambridge, New York, Melbourne, 2003.

\bibitem{plotkin70}
G.~D. Plotkin.
\newblock A further note on inductive generalization, in machine intelligence,
  volume 6, pages 101-124, 1971.

\bibitem{freqpm}
Charu~C. Aggarwal and Jiawei Han.
\newblock {\em Frequent Pattern Mining}.
\newblock Springer Publishing Company, Incorporated, 2014.

\bibitem{apriori}
Rakesh Agrawal and Ramakrishnan Srikant.
\newblock Fast algorithms for mining association rules in large databases.
\newblock In {\em Proceedings of the 20th International Conference on Very
  Large Data Bases}, VLDB '94, pages 487--499, San Francisco, CA, USA, 1994.
  Morgan Kaufmann Publishers Inc.

\bibitem{huimsurvey}
Philippe Fournier-Viger, Jerry Chun-Wei~Lin, Tin Truong-Chi, and Roger Nkambou.
\newblock {\em A Survey of High Utility Itemset Mining}, pages 1--45.
\newblock Springer International Publishing, Cham, 2019.

\bibitem{topk}
Vincent~S Tseng, Cheng-Wei Wu, Philippe Fournier-Viger, and S~Yu Philip.
\newblock Efficient algorithms for mining top-k high utility itemsets.
\newblock {\em IEEE Transactions on Knowledge and data engineering},
  28(1):54--67, 2016.

\bibitem{fold}
Farhad Shakerin, Elmer Salazar, and Gopal Gupta.
\newblock A new algorithm to automate inductive learning of default theories.
\newblock {\em {TPLP}}, 17(5-6):1010--1026, 2017.

\bibitem{uci}
M.~Lichman.
\newblock {UCI},ml repository, \url{ http://archive.ics.uci.edu/ml}, 2013.

\bibitem{aleph}
A.~Srinivasan.
\newblock {\em The Aleph Manual,
  \url{http://web.comlab.ox.ac.uk/oucl/research/areas/machlearn/Aleph/}}, 2001.

\bibitem{treeshap}
Scott~M Lundberg, Gabriel~G Erion, and Su-In Lee.
\newblock Consistent individualized feature attribution for tree ensembles.
\newblock {\em arXiv preprint arXiv:1802.03888}, 2018.

\bibitem{xgboost}
Tianqi Chen and Carlos Guestrin.
\newblock Xgboost: A scalable tree boosting system.
\newblock In {\em Proceedings of the 22Nd ACM SIGKDD}, KDD '16, pages 785--794,
  2016.

\bibitem{swiweb}
Jan Wielemaker.
\newblock {SWI-Prolog Home Page}.
\newblock \url{http://www.swi-prolog.org/}.

\bibitem{spmf}
Philippe Fournier{-}Viger, Jerry~Chun{-}Wei Lin, Antonio Gomariz, Ted Gueniche,
  Azadeh Soltani, Zhihong Deng, and Hoang~Thanh Lam.
\newblock The {SPMF} open-source data mining library version 2.
\newblock In {\em {ECML/PKDD} {(3)}}, volume 9853 of {\em LNCS}, pages 36--40.
  Springer, 2016.

\bibitem{ilp20}
Stephen Muggleton, Luc Raedt, David Poole, Ivan Bratko, Peter Flach, Katsumi
  Inoue, and Ashwin Srinivasan.
\newblock Ilp turns 20.
\newblock {\em Mach. Learn.}, 86(1):3--23, January 2012.

\bibitem{svmilp}
Stephen Muggleton, Huma Lodhi, Ata Amini, and Michael J.~E. Sternberg.
\newblock Support vector inductive logic programming.
\newblock In Achim Hoffmann, Hiroshi Motoda, and Tobias Scheffer, editors, {\em
  Discovery Science}, Berlin, Heidelberg, 2005. Springer Berlin Heidelberg.

\bibitem{kfoil}
Niels Landwehr, Andrea Passerini, Luc~De Raedt, and Paolo Frasconi.
\newblock k{FOIL}: Learning simple relational kernels.
\newblock In {\em Proceedings, The Twenty-First National Conference on
  Artificial Intelligence and the Eighteenth Innovative Applications of
  Artificial Intelligence Conference, July 16-20, 2006, Boston, Massachusetts,
  {USA}}, pages 389--394, 2006.

\bibitem{nfoil}
Niels Landwehr, Kristian Kersting, and Luc~De Raedt.
\newblock n{FOIL}: Integrating na{\"{\i}}ve bayes and {FOIL}.
\newblock In {\em Proceedings, The Twentieth National Conference on Artificial
  Intelligence and the Seventeenth Innovative Applications of Artificial
  Intelligence Conference, July 9-13, 2005, Pittsburgh, Pennsylvania, {USA}},
  pages 795--800, 2005.

\end{thebibliography}

\end{document}